\documentclass[11pt]{article}
\usepackage[utf8]{inputenc}
\usepackage{amsmath}
\usepackage{amsthm}
\usepackage{float}
\usepackage[]{algorithmic}
\usepackage{bbm}
\usepackage{verbatim}

\usepackage{algorithm}
\usepackage{adjustbox}
\usepackage{array}
\usepackage{makecell}

\usepackage{pbox}
\usepackage{kpfonts}
\usepackage{graphics}
\usepackage{multicol}
\usepackage{subfig}
\usepackage{wrapfig,lipsum,booktabs}
\usepackage[toc]{appendix}
\usepackage{authblk}
\usepackage{hyperref}
\hypersetup{
    colorlinks=true,
    linkcolor=blue,
    filecolor=magenta,      
    urlcolor=cyan,
}

\usepackage{geometry}
 \usepackage{mathptmx}

\title{Some Experiences with Hybrid Genetic Algorithms in Solving the Uncapacitated Examination Timetabling Problem}
\author[1]{Ayse Aslan}
\affil[1]{The School of Engineering and The Built Environment, Edinburgh Napier University, Edinburgh EH10 5DT, UK}
\affil[ ]{\textit{Contact: a.aslan@napier.ac.uk}}
\date{\today}

\providecommand{\keywords}[1]
{
  \small	
  \textbf{\textit{Keywords---}} #1
}

\begin{document}

\renewcommand{\algorithmiccomment}[1]{// #1}
\maketitle


\begin{abstract}
This paper provides experimental experiences on two local search hybridized genetic algorithms in solving the uncapacitated examination timetabling problem. The proposed two hybrid algorithms use partition and priority based solution representations which are inspired from successful genetic algorithms proposed for graph coloring and project scheduling problems, respectively. The algorithms use a parametrized saturation degree heuristic hybridized crossover scheme. In the experiments, the algorithms firstly are calibrated with a Design of Experiments approach and then tested on the well-known Toronto benchmark instances. The calibration shows that the hybridization prefers an intensive local search method. The experiments indicate the vitality of local search in the proposed genetic algorithms, however, experiments also show that the hybridization benefits local search as well. Interestingly, although the structures of the two algorithms are not alike, their performances are quite similar to each other and also to other state-of-the-art genetic-type algorithms proposed in the literature.


\end{abstract}

\section{Introduction}

Genetic algorithms were introduced in 1960s by Holland \cite{GAHist}. These algorithms are classified as nature inspired; specifically, they are inspired from evolutionary processes of populations. They are popularly in use to solve many complex search problems since 1990s. Being inspired by evolution, in genetic algorithms, each solution is represented as a chromosome which can be exposed to crossovers and mutations during its lifetime. Crossover operators in genetic algorithms are the mechanisms that work for intensifying search processes by favoring fit solutions and also for making these processes converge. On the other hand, mutation operators work for avoiding premature convergence by inducing some random changes in some solutions. Pure genetic algorithms (see \cite{GAinET6}), local search hybridized genetic algorithms, sometimes called \textit{memetic} algorithms (see \cite{GAinET5}), multi-objective evolutionary algorithms (MOEA) (see \cite{GAinET1} and \cite{GAinET2}) and memetic multi-objective evolutionary algorithms (see \cite{GAinET3} and \cite{GAinET4}) are used in solving various single-objective and multi-objective examination timetabling problems.

The most recent survey on examination timetabling by Qu et al. \cite{ERew} includes several studies (see \cite{GAinETs1} and \cite{GAinETs2}) which point out the issues regarding the success of genetic algorithms in solving examination timetabling problems. The common conclusion reported by those studies is that direct solution representation attempts in genetic algorithms are often not successful. Also, in the graph coloring literature the common belief is that pure genetic algorithms usually are not competitive \cite{GAGC1}. Moreover, memetic algorithms that hybridize genetic algorithms with local search methods are argued to be more successful for examination timetabling\cite{ERew}. To sum up, the common perception is that through hybridization with some local search methods and indirect solution representations genetic algorithms can provide competitive results for the examination timetabling problem. However, there are several issues regarding this approach. The first issue is on finding a suitable solution representation that is not simply direct encoding. The second issue relates to local search, about how to select a local search method to hybridize or how to balance the exploitation by local search and the exploration by the genetic mechanisms during search. For instance, an intensive local search applied on each generated solution will increase the running times of algorithms significantly, leaving less room for genetic mechanisms to produce generations.

The purpose of this paper is to provide an investigation of the success of such hybrid genetic algorithms in solving the uncapacitated examination timetabling problem. Interested readers can see \cite{EMmethod} for a recent mathematical model of this problem. For this purpose, two structurally different hybrid genetic algorithms with two different indirect solution representations are utilized. The first hybrid genetic algorithm in this study is inspired by the local search hybridized genetic algorithm developed by Galinier and Hao \cite{GAGC1} for graph coloring problems. The link between graph coloring and the uncapacitated examination timetabling is a strong one. 

The feasibility part of the uncapacitated examination problem can be represented as a conflict graph of examinations. In this graph, examinations are represented by vertices and there is an edge connecting any two vertices if there is at least one student who has to sit both of the examinations, of the connecting vertices. Differently from 

The scheduling conflicts of students are considered as the only hard constraint and the examination proximity costs are the only soft constraint considered.

Galinier and Hao's algorithm pioneered a partition based solution representation. The success of this algorithm is due to its partition based crossover, i.e., Greedy Partition Crossover (GPX). The second hybrid algorithm proposed in this study is a random key genetic algorithm that uses a priority based representation. The structure of this algorithm is inspired by the algorithm developed by Mendes et al. \cite{GA2} for a project scheduling problem. The purpose by selecting these representations is to experiment with complementary algorithms with non-similar indirect representations that will increase the depth of the investigation.

The rest of the paper is organized as follows. Section \ref{sec:2} describes the algorithms used in the paper. The computational experiments are presented in Section \ref{sec:3}. Lastly, Section \ref{sec:4} concludes the paper. 

\section{Hybrid Algorithms}\label{sec:2}

This section presents two hybrid genetic algorithms; the first one is named partition based hybrid genetic algorithm (PARHGA) and the latter is called priority based hybrid genetic algorithm (PRIHGA). The algorithms are named accordingly to their solution representation structures. Firstly, the heuristics used for generating good quality initial solutions and the two local search methods are presented. These procedures are used in both hybrid genetic algorithms.  



\subsection{Saturation Degree Heuristics} \label{sec:sat}
 A slight adaptation of the Dsatur algorithm \cite{SAT} that is used in Galinier and Hao's hybrid genetic algorithm \cite{GAGC1} for generating initial solutions is also utilized here for the same purpose, for generating good quality solutions for the initial populations of the algorithms. The adaptation made by Galinier and Hao is adding a randomization component to the Dsatur such that the algorithm can provide many diverse good quality solutions. 
 This adaptation uses random assignment for the unassigned vertices left at the end of the saturation degree based assignments. 
 
 \textit{\textbf{Saturation Degree Heuristic with Minimum Time Slot Assignment (SAT-MIN):}} This algorithm which is directly adopted from \cite{GAGC1} initially starts with an empty timetable. The algorithm uses the dynamic data on examinations of their number of time slots in the timetable in which they can still feasibly be assigned. In fact, these numbers relate to the so-called saturation degrees of vertices, with respect to the graph coloring terminology. At each iteration, the examination that has the minimum number of time slots that it can be feasibly assigned, the one with a maximum saturation degree, is selected. The description in \cite{GAGC1} does not mention what the algorithm does in tie situations in the selection phase; however, in this study ties are broken randomly. The assignment rule is such that it assigns an examination to the minimum, namely the earliest, time slot possible. When there are no more examinations left that can be feasibly assigned to some time slots, the algorithm assigns the remaining examinations to time slots in a random fashion.
 
 \textit{\textbf{Saturation Degree Heuristic with Distance Based Time Slot Assignment (SAT-DIST):}} Graph coloring and examination timetabling problems are closely related. Indeed, this relation is discovered very early and graph coloring heuristics are employed in examination timetabling frequently. However, one should be careful when directly adopting methods used for one problem to another problem. For instance, the uncapacitated examination timetabling problem that is discussed here and the graph coloring problem have completely different objectives. The first problem cares for the proximity costs of examinations, while the latter problem cares for using the least number of colors, or time slots with respect to the examination timetabling terminology. The idea in the Dsatur algorithm to assign a vertex to the minimum color possible is related to the objective of using the least number of colors, which does not exactly match with the objective of the problem discussed here. Therefore, alternatively, a distance based assignment rule is proposed in this paper. 
 
 This rule argues that as far as only the proximity costs are concerned, the prioritized examinations should be assigned to the time slots that are farthest from the center time slot. The reasoning behind this rule is that the time slots that are not close to the center will be more valuable for the examinations as they have less neighbors and therefore have less chances of incurring proximity costs. The pseudocode of the saturation degree heuristic with distance based assignment rule is given in Algorithm \ref{alg:sat}.   
 
\begin{algorithm} [H]
 \caption{Pseudocode of SAT-DIST Heuristic}
 \label{alg:sat}
 \begin{algorithmic}[1]

\STATE $k \gets \# time slots, \;\; c\gets \left \lfloor{\frac{k}{2}}\right \rfloor, \;\; e \in E\gets set\;of\;exams, \;\;Timetable \gets \emptyset$; 
\WHILE{ $\exists e \in E \;$ that can be assigned to at least one time slot feasibly }
\STATE find $e^* \in E$ that has the maximum saturation degree, break ties randomly;
\STATE find $t^* \in \{1,.., k\}$ that $e^*$ can be assigned and is farthest from $c$, break ties randomly;
\STATE assign $e^{*}$ to time slot $t^{*}$ in $Timetable$;
\ENDWHILE
\STATE for any $e\in E$ that has not been assigned yet, assign it to a time slot randomly;
\RETURN $Timetable$
\end{algorithmic}
\end{algorithm}



\subsection{Local Search Methods}
The genetic algorithms proposed here are hybridized with local search. This hybridization applies local search on the produced offspring solutions by crossover mechanisms. This paper presents two local search methods to use for this purpose. The first one is a computationally light method, while the second one is much intensive. In hybridizing genetic algorithms with local search methods the issue is on balancing the exploitation by local search. A less intensive local search would be more time efficient, leading genetic algorithms to produce many generations, while an intensive local search method can improve the quality of offspring solutions better.



\subsubsection{Vertex Descent Local Search (VDLS)}
The hybrid algorithm by Galinier and Hao \cite{GAGC1} for graph coloring problems uses a tabu search method. A following study \cite{GAGC1follow} examines the components of Galinier and Hao's algorithm to understand what makes this algorithm successful. The authors of the following study conclude that the tabu search is not really the essential part; in fact, the authors replace the tabu search with a much simpler vertex descent method and they can however obtain similar results. Following that, the computationally light vertex descent method is also considered here. In the implementation of this method in this study, the efficient data structures described in this following study on the relative costs of the assignment of examinations to time slots are used as well. The pseudocode of this method is given in Algorithm \ref{alg:vd}.

\begin{algorithm} [H]
 \caption{Pseudocode of VDLS}
 \label{alg:vd}
 \begin{algorithmic}[1]

\STATE $k \gets \# time slots,\;\; e \in E\gets set\;of\;exams, \;\; Timetable[e \in E] \in \{1,...,k\}$; 
\WHILE{an improvement can be made in the cost function}
\FOR{ $e \in E$}
\STATE assign $e$ to $t^* \in \{1,..,k\}$ in $Timetable$ that is least costly for the $Timetable$
\ENDFOR
\ENDWHILE
\RETURN $Timetable$
\end{algorithmic}
\end{algorithm}


\subsubsection{Hyper-Heuristic Local Search (HHLS)}

Hyper-heuristics are popularly in use for many educational timetabling problems in general \cite{HH}. A selection type perturbation hyper-heuristic framework is presented here as an intensive local search tool. These types of hyper-heuristics briefly consist of three components: a pre-determined pool of low-level heuristics, a high-level heuristic selection mechanism and a move-acceptance criterion. They perform local search on a single-solution by iteratively selecting one of the low-level heuristics and then applying it on the current solution to obtain a new solution, and lastly the move-acceptance criterion decides if the new solution is going to be accepted as the current solution or not \cite{ClassificationHH}.  

The framework presented in this paper consists of five low-level heuristics, which most of them are taken from the operators used in a recent variable neighbourhood search study \cite{VNS} of examination timetabling problems. These heuristics/operators are briefly described below.   




\begin{itemize}
    \item $LLH_{1}$: Moves a randomly selected examination to a randomly picked time slot feasibly. 
    \item $LLH_{2}$: Selects an examination randomly that can feasibly be moved to at least one different time slot, then finds a second examination that can feasibly be swapped with the first examination such that their swap will reduce costs best.
    \item $LLH_{3}$: This operator named \textit{Kempe Chain}, having origins in the graph coloring literature, was first shown in \cite{Kempe} to be useful for the examination timetabling as well.  This heuristic randomly selects a starting examination and a time slot that is different than the assigned slot to the starting examination, then it swaps a first group of examinations that are assigned to the picked time slot and in conflict with the starting examination and a second group of examinations that are assigned to the time slot of the starting examination and in conflict with the first group of examinations. The main aim of this operator is to increase the flexibility in timetables.
    \item $LLH_{4}$: Randomly selects a time slot and randomly selects another time slot to move the examinations of the first slot. The times of the examinations after the secondly selected time slot are moved forwarded. 
    \item $LLH_{5}$: Selects randomly two time slots and swaps the examinations assigned to them.
\end{itemize}

The framework here uses simple random selection mechanism that selects low-level heuristics uniformly and accepts solutions that are non-worsening at each iteration. 

The pseudocode of the hyper-heuristic framework is given in Algorithm \ref{alg:hh}. The issue with this local search method is the determination of its number of iterations. In this paper, this parameter is tuned offline to 25,000 iterations. Also, a second parameter is used for terminating the local search when no improvement is observed for a long time. This parameter is also offline tuned and fixed to 5,000 iterations in all experiments. 

Moreover, the technique of applying local search only on some produced offspring solutions is also experimented. This technique is useful to reduce the computational burden of the intensive local search and therefore it lets algorithms produce more generations. However, it is observed that with this technique the algorithms proposed here perform worse. Therefore, it is decided not to make use of this technique.    

\begin{algorithm} [H]
 \caption{Pseudocode of HHLS}
 \label{alg:hh}
 \begin{algorithmic}[1]

\STATE $Pool \gets \{ LLH_1, \ldots, LLH_{5}\}, s \gets Timetable $; 
\STATE $f_{s} \gets Calculate\;Objective(s)$; 
\WHILE{ (iteration limit \& non-improvement limit) not reached}
\STATE $h \gets $ randomly select a low-level heuristic from $Pool$
\STATE $s_{new} \gets Apply (s, LLH_{h})$;
\STATE $f_{new} \gets Calculate\;Objective(s_{new})$;
\IF{$f_{new} \leq f_{s}$}
\STATE $s \gets s_{new}$;
\STATE $f_{s} \gets f_{new}$;
\ENDIF
\ENDWHILE
\RETURN $s$
\end{algorithmic}
\end{algorithm}

\subsection{Partition Based Hybrid Genetic Algorithm (PARHGA)}

The structure of the partition based hybrid genetic algorithm that this paper proposes is very similar to Galinier and Hao's \cite{GAGC1} hybrid algorithm for graph coloring problems. In this algorithm, each solution $s_a$ is represented by a set of $k$ many partition sets $\{V_1^a,..., V_{k}^a\}$ which consist of examinations, where $k$ is the number of time slots. The generation update in this algorithm selects two parent solutions to produce only one offspring each time. This algorithm replaces the mutation phase of traditional genetic algorithms with local search. Namely, after crossover operator is used to generate an offspring from two parents, the resulting solution is enhanced in quality with local search instead of being exposed to some random changes that will increase the diversity of the solution. The population update rule applied in this algorithm again favors increasing the quality of solutions in the population; it replaces the worst quality parent with the local search enhanced offspring. However, parent selection phase is not elitist; at each iteration two distinct parents are randomly selected. The Algorithm \ref{alg:ga1} presents the pseudocode of this hybrid genetic algorithm. 

\begin{algorithm} [H]
 \caption{Pseudocode of PARHGA}
 \label{alg:ga1}
 \begin{algorithmic}[1]

\STATE $n\gets$ population size, $P\gets \emptyset$ ; 
\FOR{$i \in \{1,...,n\}$}
\STATE $s_i \gets GenerateSolution()$; 
\STATE $s_i \gets LocalSearch(s_i)$;
\STATE $P\gets P \cup \{ s_i\}$;
\ENDFOR
\WHILE{ time limit not reached}
\STATE select randomly $s_a \in P$ and $s_b \neq s_a \in P$;
\STATE $s_{new} \gets HybridPartitionCrossover(s_a,s_b)$;
\STATE $s_{new}\gets LocalSearch(s_{new})$;
\STATE $P\gets P \cup \{s_{new}\}$;
\IF{$Fitness_{s_{a}} < Fitness_{s_{b}}$}
\STATE $P \gets P - \{s_a\}$;
\ELSE
\STATE $P \gets P - \{s_b\}$;
\ENDIF
\ENDWHILE
\RETURN $P$
\end{algorithmic}
\end{algorithm}

The success of Galinier and Hao's algorithm in the graph coloring problem is mainly due to its greedy partition crossover (GPX). The remark made by Galinier and Hao on direct representation approaches, which explicitly assign each vertex to a color, is that labelling of vertices is not actually providing any helpful information, instead the partition of vertices into independent sets is the valuable information in essence. GPX crossover realizes the importance of these independent sets and operates on the partitions of vertices. In particular, large independent sets are the most useful ones. These sets provide the most efficient ways of using a color, using the same color to color a large number of vertices. That's why GPX aims to make use of large independent sets of both parent solutions. GPX alternately considers parents and each time transfers the partition set with the maximum cardinality to offspring.      

This study considers a more greedy version of GPX, instead of alternating between two parents, both parents are considered during every iteration of the crossover. Additionally, the crossover proposed in this study is a parameterized hybrid crossover. The hybridization is with the saturation degree heuristic that is described in Section \ref{sec:sat} and the level of this hybridization is parameterized ($r \in [0,1]$). The reasoning behind this parametrization is to experiment with different values of hybridization. The main idea behind this hybridization is to lighten the crossover in a way to increase the chances of producing feasible offspring solutions. The observation made with GPX crossover is that the first set of iterations of the crossover are the most useful, letting the transmission of large non-conflicting partition sets of parents to offspring. However, the transmitted partitions get smaller and smaller in time, not giving much valuable information by the end, and after some point the transmitted partitions may cause more harm than help by decreasing the flexibility of the timetable being produced. The saturation degree heuristic that is known to be successful for producing feasible solutions has the potential to be a promising tool for completing the partially built timetables. The saturation degree heuristic hybridized greedy partition crossover (SATHGPX) presented here applies crossover up to some point to obtain a partial timetable and then it uses a saturation degree heuristic to complete the timetable. The Algorithm in \ref{alg:cx} describes SATHGPX. This crossover can use either of the saturation degree heuristics.  
    
\begin{algorithm} [H]
 \caption{Pseudocode of SATHGPX}
 \label{alg:cx}
 \begin{algorithmic}[1]
\STATE $k\gets \#timeslots,\; r \in [0,1],\; s_{a}\gets \{V_{1}^{a},...,V_{k}^{a}\},\; s_{b} \gets \{V_{1}^{b},...,V_{k}^{b}\}$; 
\STATE $s_{new} \gets \{V_{1}^{new} \gets \emptyset, ...,V_{k}^{new} \gets \emptyset\}$;
\FOR{$i \in \{1,...,\left \lfloor rk \right \rfloor \}$}
\STATE select $j^{*}\in \{1,...,k\}$ and $m^{*}\in\{a,b\}$ such that $V_{j^{*}}^{m^{*}}$ has the maximum cardinality; 
\STATE $V_{i}^{new} \gets V_{j^{*}}^{m^{*}} $;
\STATE remove vertices of $V_{i}^{new}$ from $s_{a}$ and $s_{b}$; 
\ENDFOR
\STATE Apply SAT-MIN or SAT-DIST heuristic to assign unassigned exams of $s_{new}$;
\RETURN $s_{new}$
\end{algorithmic}
\end{algorithm}

The cost of a solution is calculated in the following way in both of the hybrid algorithms. The examination conflicts in the solution are counted and penalized with a large weight value and this is added to the proximity cost of the solution to compose the final cost of the solution.

\subsection{Priority Based Hybrid Genetic Algorithm (PRIHGA)} 

The genetic characteristics of PRIHGA are similar to the random key genetic algorithm in \cite{GA2} that is proposed for a project scheduling problem. In a way, PRIHGA can be considered as the local search hybridized version of Mendes et al.'s algorithm \cite{GA2}. This algorithm represents a solution $s_a$ with the priorities of examinations $\{\pi_1^a,..., \pi_m^a\}$ where $\pi_i^a \in [0,1], \forall i$ and $m$ is the number of examinations.

At each generation, $n_{sel}$ many solutions with the best fitness values, the top quality solutions, in a generation are transmitted to the population of the next generation for the sake of elitism. At each generation, $n_{cross}$ many crossovers are performed. The applied crossover selects two parent solutions and generates only one offspring from them. Specifically, the applied crossover here is a biased uniform crossover. The parent selection in the crossover phase is partly elitist; one of the parents is chosen among the top quality solutions. The uniform crossover generates an offspring by gene-by-gene transmission. In each transmission, a biased coin is tossed to select the parent of the gene; the bias reflects the elitist probability $p_{elit}$ of favoring the top quality solution. All generated offspring solutions are applied local search for the enhancement of their qualities. At each generation, a set of randomly generated solutions, $n_{mig}$ many, are migrated to replace the solutions that are not among the best. It is chosen here to not to apply local search on these migration solutions for the sake of increasing the diversity of the genes in the population. The pseudocode in Algorithm \ref{alg:ga2} gives the details of this algorithm. 

This random key algorithm uses the priority based representation of each examination with a continuous value in $[0,1]$. The direct solution and its fitness value can be obtained via a decoding procedure. The decoding procedure used in this algorithm is the simple assignment rule that orders the exams by their priorities and assigns them to the earliest time slots possible, similar to the assignment used in SAT-MIN heuristic. 



\begin{algorithm} [H]
 \caption{Pseudocode of PRIHGA}
 \label{alg:ga2}
 \begin{algorithmic}[1]

\STATE $P\gets \emptyset$ ; 
\STATE Initialize $ n_{sel}, n_{cross}, n_{mig}, n= n_{sel}+n_{cross}+n_{mig}, p_{elit}$;  
\FOR{$i \in \{1,...,n\}$}
\STATE $s_i \gets GenerateSolution()$; 
\STATE $s_i \gets LocalSearch(s_i)$;
\STATE $P\gets P \cup \{ s_i\}$;
\ENDFOR
\WHILE{ time limit not reached}
\STATE $P_{top} \gets SelectTopSolutions(P, n_{sel}), P_{cx} \gets \emptyset, P_{mig} \gets \emptyset$;
\FOR{$i \in \{1,..,n_{cross}\}$}
\STATE select randomly $s_a \in P_{top}$ and $s_b \in P-P_{top}$;
\STATE $s_{new} \gets HybridUniformCrossover(p_{elit},s_a,s_b)$;
\STATE $s_{new} \gets LocalSearch(s_{new})$;
\STATE $P_{cx}\gets P_{cx} \cup \{s_{new}\}$;
\ENDFOR
 \STATE $P_{mig}\gets GenerateSolution(n_{mig})$;
\STATE $P \gets P_{top} \cup P_{cx} \cup P_{mig}$;
\ENDWHILE
\RETURN $P$
\end{algorithmic}
\end{algorithm}

This study again considers a parameterized saturation degree heuristic hybridized crossover scheme, this time for the biased uniform crossover (SATHUCX). Again, the main purpose with this hybridization is to lighten the crossover and utilize saturation degree heuristic to increase the chances of producing feasible offspring solutions. 

\begin{algorithm} [H]
 \caption{Pseudocode of SATHUCX}
 \label{alg:cx2}
 \begin{algorithmic}[1]
\STATE $m\gets \#exams, l \gets \{1,.., m\}, \; r \in [0,1],\; s_{a}\gets \{\pi_{1}^{a},...,\pi_{m}^{a}\},\; s_{b} \gets \{\pi_{1}^{b},...,\pi_{m}^{b}\}$; 
\STATE $s_{new} \gets \{\pi_{1}^{new} \gets 0, ...,\pi_{m}^{new} \gets 0 \}$;
\FOR{$i \in \{1,...,\left \lfloor rm \right \rfloor \}$}
\STATE $p \gets SelectParent(p_{elit},a,b)$;
\STATE select $j^{*}\in l$ such that $\pi_{j^{*}}^{p} \geq \pi_{k}^{p}, \forall k \in l $;
\STATE $\pi_{j^{*}}^{new} \gets \pi_{j^{*}}^{k}$;
\STATE remove $j^{*}$ from $l$; 
\ENDFOR
\STATE Apply SAT-MIN or SAT-DIST heuristic to assign priorities to the unassigned examinations of $s_{new}$;
\RETURN $s_{new}$
\end{algorithmic}
\end{algorithm}

The pseudocode of the uniform hybrid crossover is given in Algorithm \ref{alg:cx2}. The crossover is only applied to some extent, based on the value of $r \in [0,1]$. The idea is to transmit the most valuable genes of the parents during the crossover phase. Here it is suggested that the genes informing of the high-importance of examinations can be more useful. Following that, the examinations that have the highest priority levels in the parent solutions are transmitted during crossover first. The partially built solution by the biased uniform crossover is then completed by one of the saturation degree heuristics.

\section{Computational Experiments}\label{sec:3}

The experiments are conducted on the Toronto benchmark instances of the uncapacitated examination timetabling. The characteristics of the instances are given in Table \ref{tab:inst}. Both algorithms are implemented in C++ language and all experiments are conducted using a cluster of computing nodes with Intel Xeon CPU 2.5 GHz processors. In all statistical tests, 95 percent confidence interval is considered.    

The examination proximity costs of students are the only considered objective in these benchmark instances. The number of time slots are already given in the instances; the objective is not of the reduction of time slots to use. A penalty of $w^{s}=2^{(5-s)}$ is incurred each time a student has to sit two examinations that are only $s \in \{1,2,3,4,5\}$ slots apart. The total of these penalties are averaged over the total number of students and this average is reported as the objective value of the problem.  

\begin{table}[H]
    \centering
    \begin{tabular}{l|l|l|l|l}
    Instances & \#Exams & \#Students & \#Time slots & Density of the\\ 
     & & & & conflict matrix\\ 
    \hline 
        Car-f-92 & 543& 18,419 & 32& 0.14 \\
        Car-s-91 & 682 & 16,925 & 35 & 0.13 \\
        Ear-f-83 & 190 & 1,125 & 24& 0.27 \\
        Hec-s-92 & 81 & 2,823 & 18 & 0.42 \\
        Kfu-s-93 & 461 & 5,349 & 20& 0.06 \\
        Lse-f-91 & 381 & 2,726 & 18& 0.06 \\
        Pur-s-93 & 2,419 & 30,032 & 43 & 0.03 \\
        Rye-s-93 & 486 & 11,483 & 23& 0.07 \\
        Sta-f-83 & 139 & 611& 13& 0.14 \\
        Tre-s-92 & 261 & 4360 & 23 & 0.18 \\
        Uta-s-92 & 622 & 21,267 & 35 & 0.13 \\
        Ute-s-92 & 184 & 2,750 & 10 & 0.08 \\
        Yor-f-83 & 181 &941& 21 & 0.29 \\
        \hline
    \end{tabular}
    \caption{Toronto Benchmark Instances}
    \label{tab:inst}
\end{table}

\subsection{Saturation Degree Heuristic: Assignment Rule Makes a Difference}


Two saturation degree heuristics, SAT-MIN and SAT-DIST, are tested on the Toronto instances to investigate if a distance based assignment provides significant benefits or not. For each instance, 50 runs are completed, and in each run 100 samples are taken by the heuristics. The best quality solutions and the number of feasible solutions found in the runs are recorded and used for the comparison of the assignment rules. The averages of the best solutions found in the runs are given in the first columns of the corresponding heuristics in Table \ref{tab:SATsols} while the second columns in Table \ref{tab:SATsols} give the averages of the number of feasible solutions found in the runs with 100 samples. One-way ANOVA and the non-parametric Mann-Whitney U test are used to detect significant differences in the performances of the heuristics. Both detection approaches reveal that in all instances SAT-DIST significantly performs better than SAT-MIN in terms of the quality of the best average solutions found in the runs, while no significance is found in terms of the quality with respect to producing feasible solutions. However, the average results indicate that SAT-MIN is not worse than SAT-DIST in terms of producing feasible solutions. This could be directly related to the issue that the graph coloring objective, minimizing the number of colors used, that is reflected in SAT-MIN assignment performs good with respect to the feasibility of the examination timetabling problem which uses a finite number of time slots. The superior performance of the distance based assignment over the minimum assignment rule in SAT-MIN shows that there is a potential in improving the performances of the graph coloring heuristics for the examination timetabling, even through some small changes.


\begin{table}[H]
    \centering
    \begin{tabular}{l|l l|l l}
    \hline
    & SAT-MIN & & SAT-DIST & \\
    Data sets & BestSol (ave) & \#FeasSols (ave) & BestSol(ave) & \#FeasSols (ave)\\
    \hline 
        Car-f-92 & 9.16 & \textbf{99.42} & \textbf{8.26} & 99.38 \\
        Car-s-91 & 11.11 & \textbf{99.98} & \textbf{10.08} & 99.96 \\
        Ear-f-83 & 60.65 &  \textbf{85.42} & \textbf{55.41} &  84.12\\
        Hec-s-92 & 19.05 & 50 & \textbf{17.02} & \textbf{51.48} \\
        Kfu-s-93 &  34.71 & \textbf{70.48} & \textbf{28.70} & 69.22 \\
        Lse-f-91 &  25.87 & \textbf{72.28} & \textbf{21.23} & 70.84\\
        Pur-s-93 &  15.64 & 99.98 & \textbf{13.8} & \textbf{100}\\
        Rye-s-93 &  25.49 & 61.12 & \textbf{21.28} & \textbf{61.54} \\
        Sta-f-83 &  184.90 & \textbf{100} & \textbf{165.79} & 99.96 \\
        Tre-s-92 &  14.62 & \textbf{87.64} & \textbf{13.13} & 87.2 \\
        Uta-s-92 &  7.08 & \textbf{99.56} & \textbf{6.42} & 99.52 \\
        Ute-s-92 &  47.83 & \textbf{56.86} & \textbf{36.78} & 56.54 \\
        Yor-f-83 &  56.93 & \textbf{36.82} & \textbf{51.93} & 35.22 \\
        \hline
    \end{tabular}
    \caption{SAT-MIN versus SAT-DIST}
    \label{tab:SATsols}
\end{table}


\subsection{Calibrating PARHGA}
\label{sec:cal_1}

Both algorithms are calibrated based on a Design of Experiments approach \cite{GAexper}. The purpose of these experiments is to fine tune the algorithms before benchmarking and also to understand the important factors affecting the performances of the algorithms. For the first algorithm, PARHGA, the following parameter settings are cross experimented.

\begin{itemize}
    \item local search method: VDLS only and VDLS + HHLS,
    \item population size (n): 20, 50 and 100,
    \item heuristic solution percentage in initial population: 50\% and 100\%,
    \item initial solution heuristic: SAT-MIN and SAT-DIST,
    \item SAT hybridization percentage level in crossover (100(1-r)): 0\%, 25\%, 50\% and 75\%,
\end{itemize}

The total number of experiments per instance is 96. In order to better understand the differences in the performances of the algorithm with respect to the parameter settings tested, two hours long runs are considered. For the purposes of practicality, only two out of 13 benchmark instances are selected for the calibration experiments. The selected instances are Yor-f-83 and Lse-f-91. Lse-f-91 instance has a medium size with respect to the number of exams, while Yor-f-83 is relatively smaller but it is denser, compared to other instances. The performance metric used for measuring the quality of the solutions found by the algorithm is the distance to the best found solutions in the literature. This is to provide a common performance metric for two different instances. Specifically, the relative percentage deviation from the best solution is measured. Relative Percentage Deviation (RPD) of a solution $S$ with value $Val(S)$ for an instance $I$ with best known solution value $BestSol_{I}$ is $\frac{Val(S)- BestSol_{I}}{BestSol_{I}} \times 100$.

One-way ANOVA is sometimes used to detect the effects of varying parameters of algorithms. For instance, in \cite{GAexper} authors use the one-step F-value approach with one-way ANOVA to calibrate and also to detect the significance of factors of their genetic algorithm. The issue with ANOVA is that there are several assumptions that need to hold for ANOVA results to be valid such as the normality assumption of the distributions. Here, the non-parametric approach of multi-layer perceptron is also utilized along with ANOVA for identifying the significance levels of the factors. However, the calibration of the hybrid genetic algorithm is done by the one-step F-value approach. This approach firstly finds the single factor that has the highest F-value with respect to the RPD values and then it fixes the factor to the setting that has the lowest RPD mean. Then, the samples are reduced to the ones that are of the fixed setting. One by one, each factor is fixed in this way.  

\begin{figure}[H]
    \centering
    \subfloat[Local Search Method Factor]{{\includegraphics[scale=0.5]{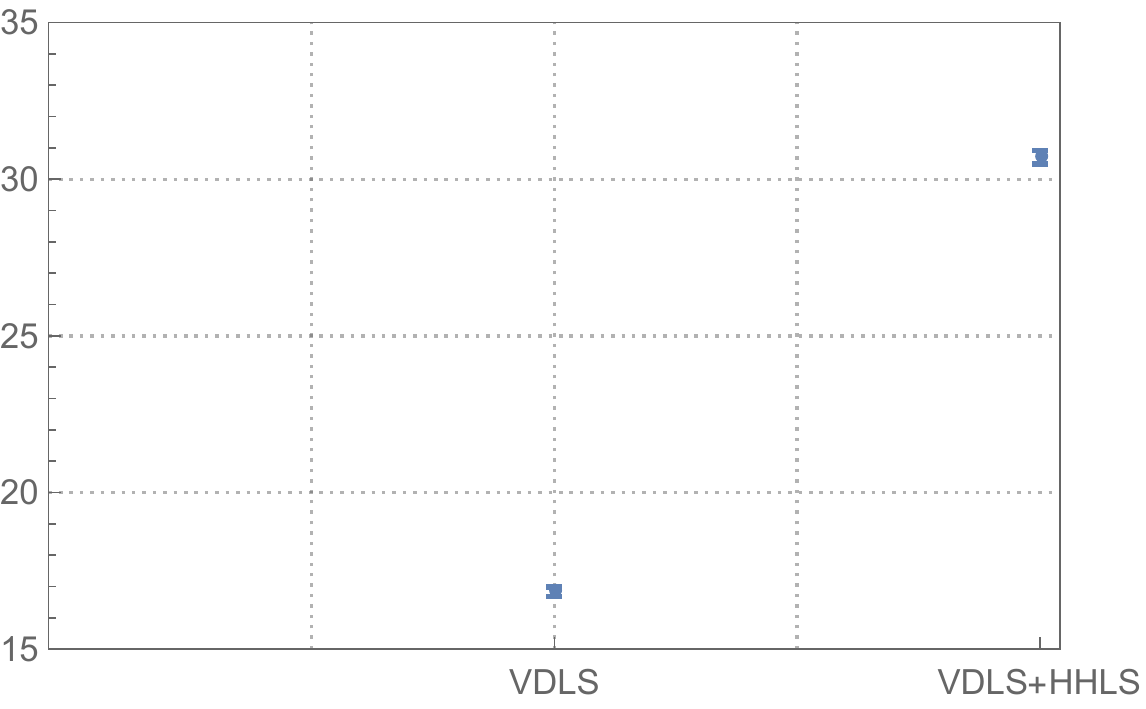} }}%
    \qquad
    \subfloat[SAT Hybridization Percentage Level in Crossover Factor]{{\includegraphics[scale=0.5]{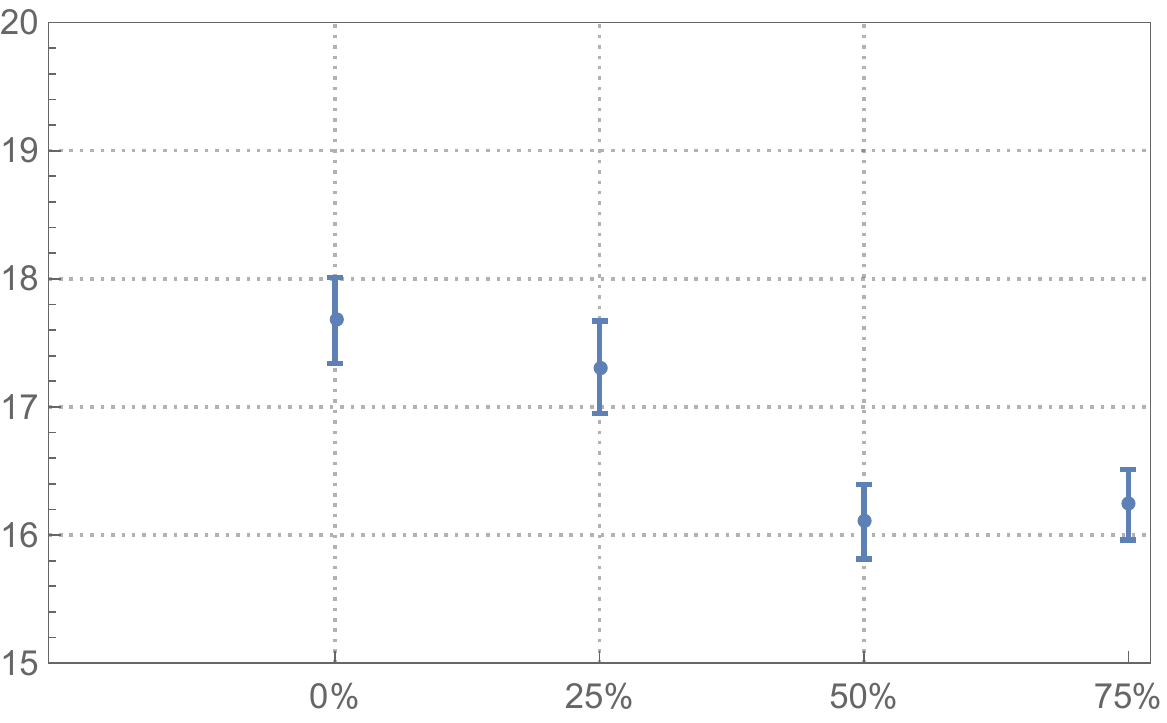} }}%
  \caption{Important Factors of PARHGA with respect to RPD}
    \label{fig:fac_g1}
\end{figure}
 
The Figure \ref{fig:fac_g1} shows the error bounds of the means, with 95\% confidence levels, of the two factors that are shown statistically significant for PARHGA, with respect to their p-values in one-way ANOVA. 
 
The most important factor, by far, is found to be the local search method with ANOVA (F-value=2747.03). VDLS is a lot quicker than HHLS and therefore the algorithm that only uses VDLS is able to produce a lot more generations than the one that uses both local search methods under a time limit. Approximately, PARHGA with only VDLS is able produce around 30-40 times more generations than the one with intensive local search, in both instances tested. However, the experiments do not show indications that the efficiency advantage of only using VDLS is also useful for producing good quality solutions. The RPD performance of the algorithm that uses both local search methods is significantly better. 

The second most significant factor is the SAT hybridization percentage in crossover (F-value=6.014). The means of the RPD values are better when the SAT hybridization level is higher ($\geq 50\%$). This indicates that SAT completion of a solution that is partially built with crossover can be better than the solution that is built by full crossover, the case where $r=1$.     

The remaining factors are not found to be significant (p-values $\leq$ 0.05). However, they are tuned accordingly to the one-step F-value approach; initial population is tuned to 20, initial solution heuristic is tuned to SAT-DIST and lastly 50\% of the initial population is tuned to be produced by heuristic solutions.   

The solutions produced by the saturation degree heuristics are applied local search (only VDLS in the experiments) before they are passed as the initial population of the both hybrid algorithms. According to observations, although SAT-DIST is significantly superior to SAT-MIN, after the local search phase the quality of the solutions by the two heuristics are very similar. This could be the reason why the initial solution heuristic factor is not significant for PARHGA.

\begin{figure}[H]
    \centering
    \subfloat[SAT-MIN]{{\includegraphics[scale=0.5]{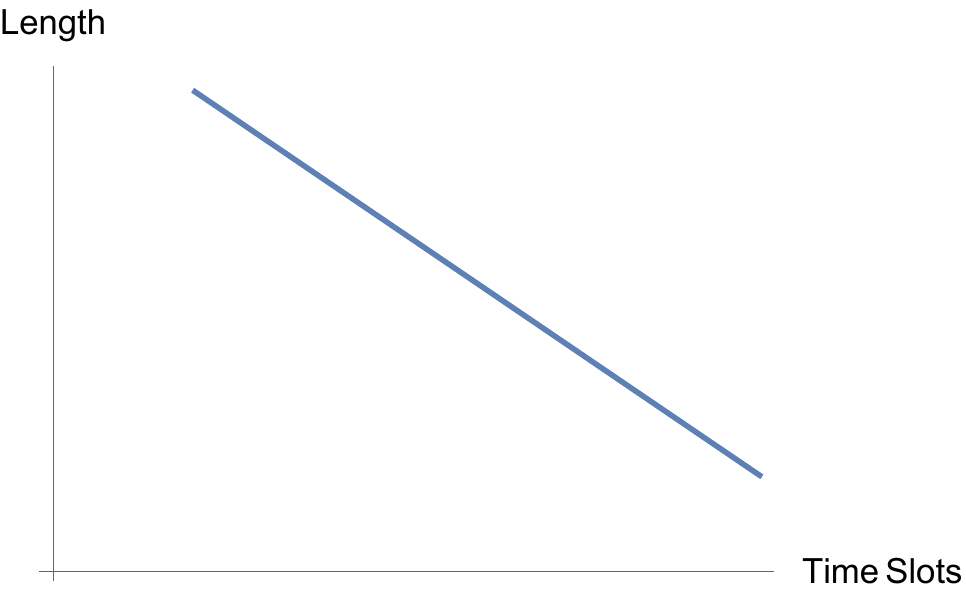}}} %
    \qquad
    \subfloat[SAT-DIST]{{\includegraphics[scale=0.5]{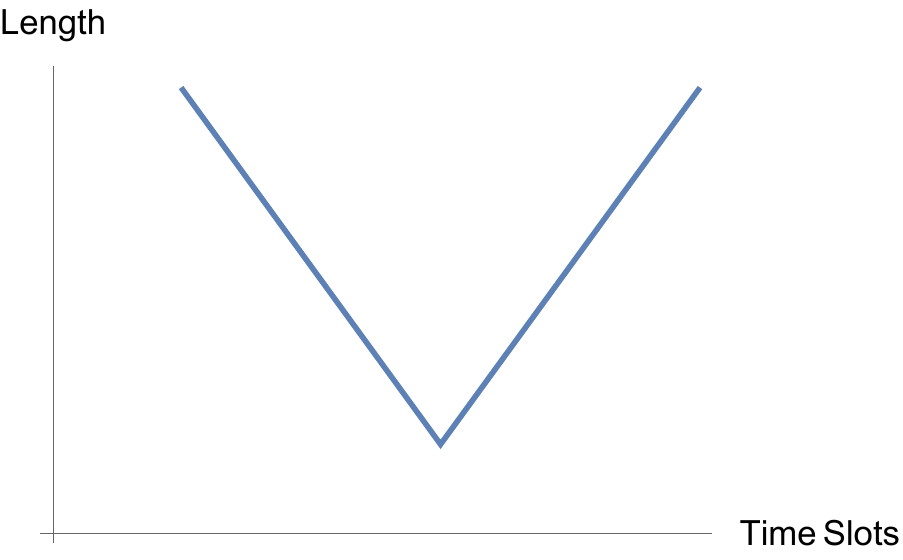}}} %
     \qquad
    \subfloat[After Local Search on SAT-MIN or SAT-DIST]{{\includegraphics[scale=0.5]{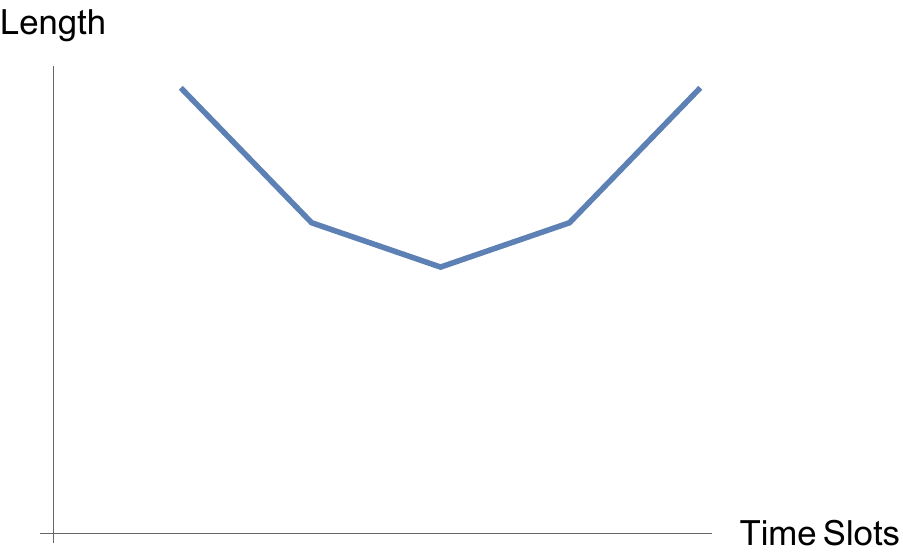}}} %
  \caption{The Outlooks of Time Slots by SAT Heuristics}
    \label{fig:sat}
\end{figure}

The Figure \ref{fig:sat} illustrates the outlook of the time slots, in terms their lengths (the number of examinations assigned to them), of the solutions produced by SAT-MIN and SAT-DIST respectively and after the local search applied on these solutions. The outlooks of the SAT-MIN and SAT-DIST solutions are directly related to their assignment rules; for instance, SAT-MIN firstly fills the earlier time slots. Note that the outlook after the local search gets very similar for both SAT-MIN and SAT-DIST solutions and in fact this outlook is more close to the outlook of the SAT-DIST solutions. 

\begin{figure}[H]
    \centering
    \includegraphics[scale=0.5]{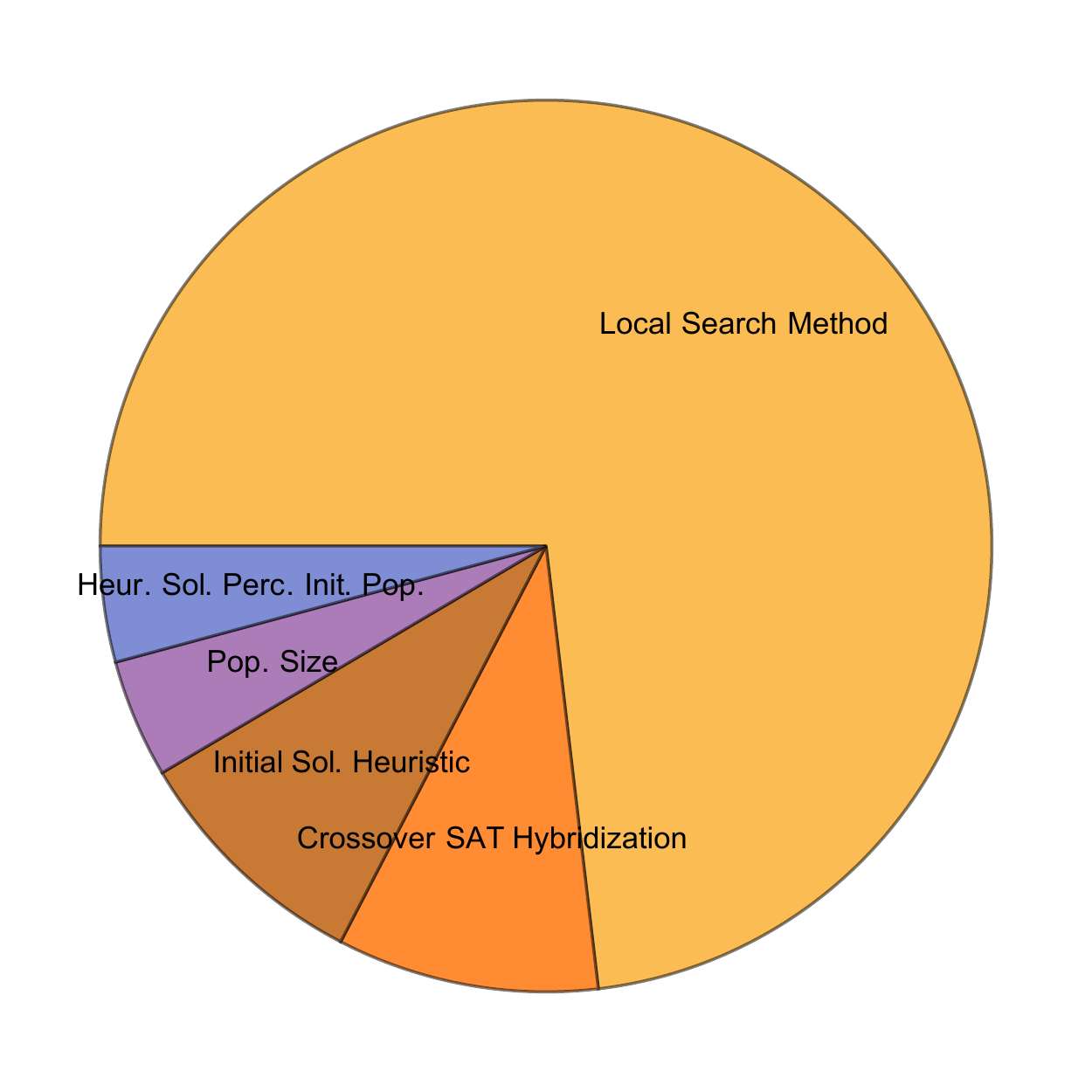}
    \caption{Pie Chart Output of the Perceptron Factor Importance Analysis for PARHGA}
    \label{fig:percept12}
\end{figure}

The importance levels of the factors are also measured with the help of a multi-layer perceptron. The multi-layer perceptron tool in SPSS 25 is utilized for this purpose. The standard model architecture setting of the tool is used that takes 70\% of the samples as training and the remaining for testing. The perceptron sensitivity analysis tool of independent variable importance analysis is used for obtaining a second input on the importance levels of the factors, along with ANOVA. The output of this analysis provides percentage importance levels of each factor with respect to their sensitivity to the perceptron model being produced for predicting the RPD values. This output gives insights into which factors are playing important roles in explaining the RPD measures of the samples. However, the output of this analysis for the factors of PARHGA is totally in line with ANOVA approach. The analysis by the perceptron model also indicates that the local search method is by far the most significant factor and SAT hybridization level in crossover is the second most important factor. The Figure \ref{fig:percept12} shows the importance levels of the factors in a pie chart.  


\subsection{Calibrating PRIHGA}
\label{sec:Cal2}

After the intensive local search is shown significantly superior in the calibration experiments of PARHGA, PRIHGA is also directly calibrated to VDLS+HHLS local search method. The parameter settings experimented for calibrating PRIHGA are as follows.

\begin{itemize}
    \item population size ($n$): 20, 50 and 100,
    \item percentages of selection and migration over the population (to determine $n_{sel}$ and $ n_{mig}$ in Algorithm \ref{alg:ga2}): 10\% selection and 10\% migration, 20\% selection and 20\% migration, and 25\% selection and 25\% migration,
    \item elitism probability($p_{elit}$): 0.6 and 0.8,
    \item SAT hybridization percentage level in crossover (100(1-r)): 0\%, 25\%, 50\% and 75\%,
    
\end{itemize}

Since with the local search methods there is no real difference between SAT-MIN and SAT-DIST heuristics, here for PRIHGA, the initial solution generating heuristic is fixed to one of them, specifically to SAT-MIN. PRIHGA, differently from the first hybrid algorithm, contains a migration phase that provides randomly generated solutions to each generation. For this reason, experimenting on the fraction of the heuristic solutions in the initial population does not seem worthwhile and therefore all solutions of the initial population are fixed to be generated by the heuristic.   

The calibration approach is the same as in the first algorithm, the one-step F-value approach. Again Yor-f-83 and Lse-f-91 instances are selected in the experiments and RPD measure is used to compare different parameter settings. 

\begin{figure}[H]
    \centering
    \subfloat[Population Size]{{\includegraphics[scale=0.5]{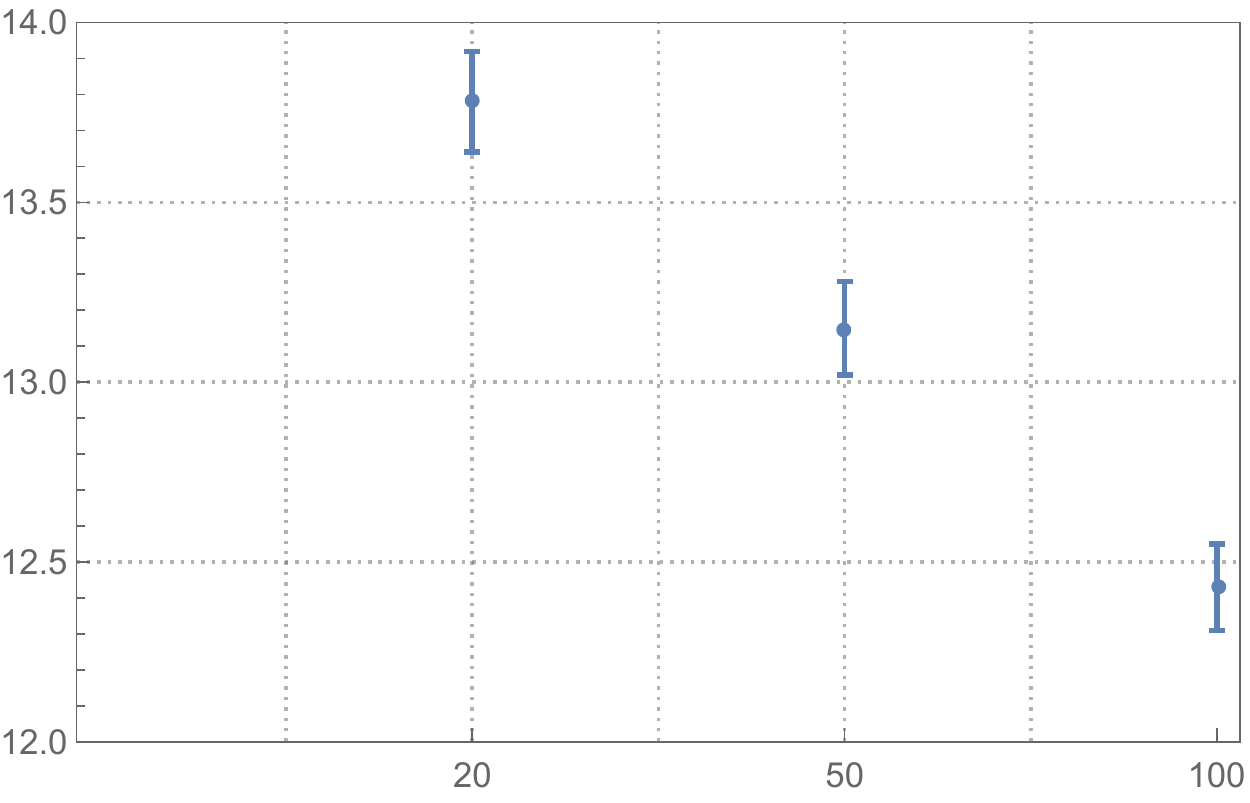}}} %
    \qquad
    \subfloat[SAT Hybridization Percentage Level in Crossover]{{\includegraphics[scale=0.5]{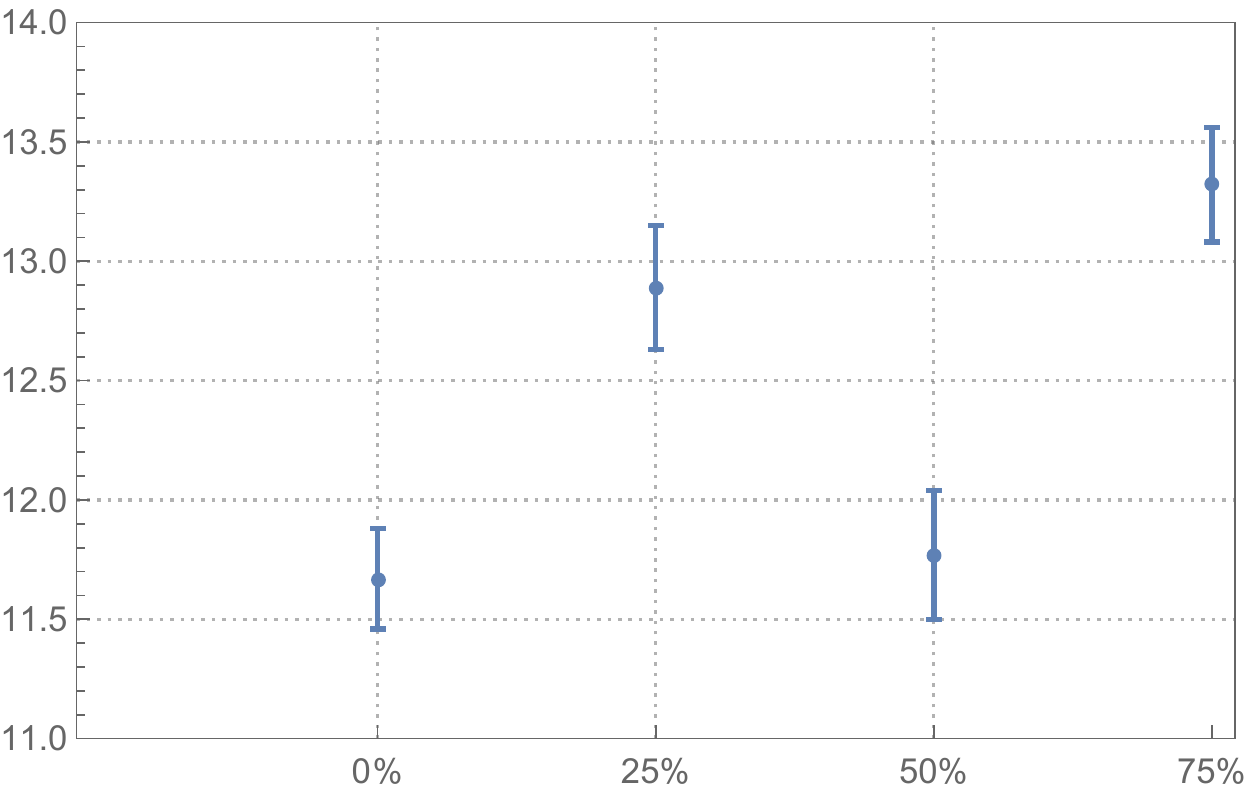}}} %
     \qquad
    \subfloat[Elitist Probability]{{\includegraphics[scale=0.5]{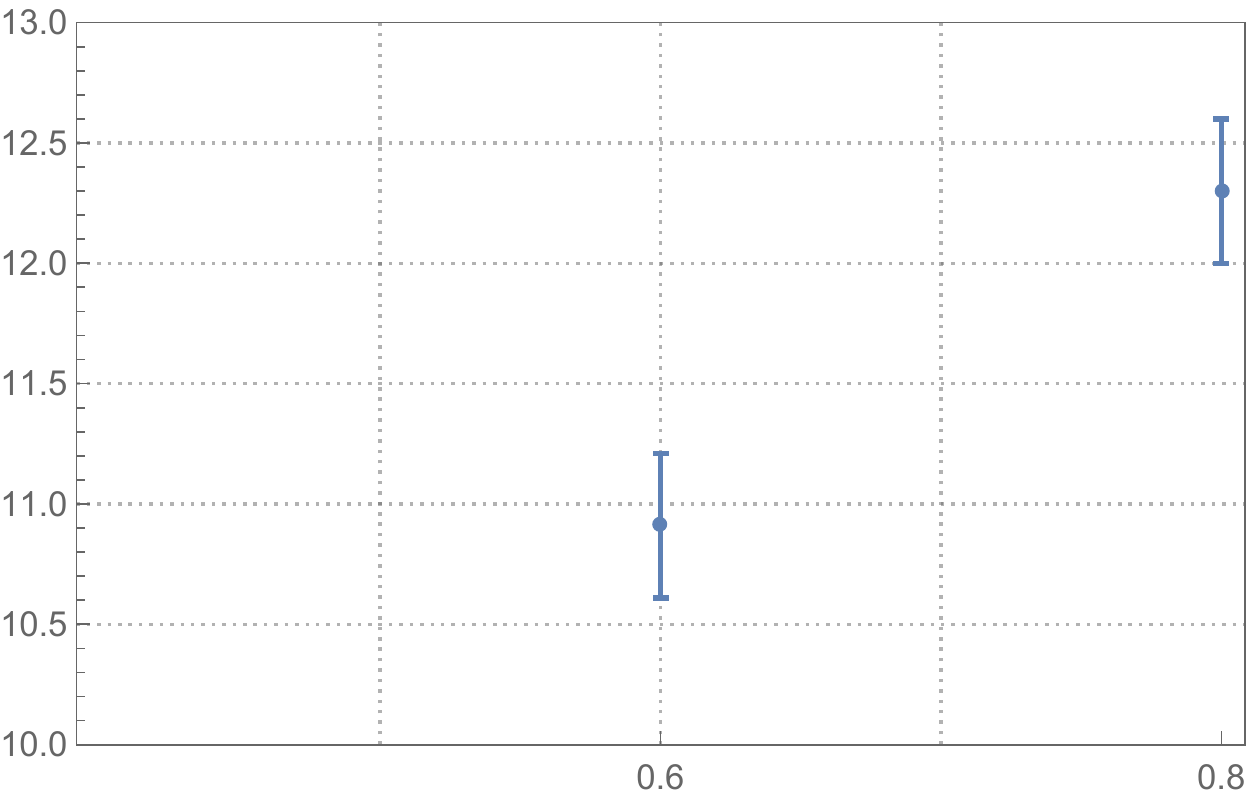}}} %
     \qquad
    \subfloat[Selection and Migration Percentages]{{\includegraphics[scale=0.5]{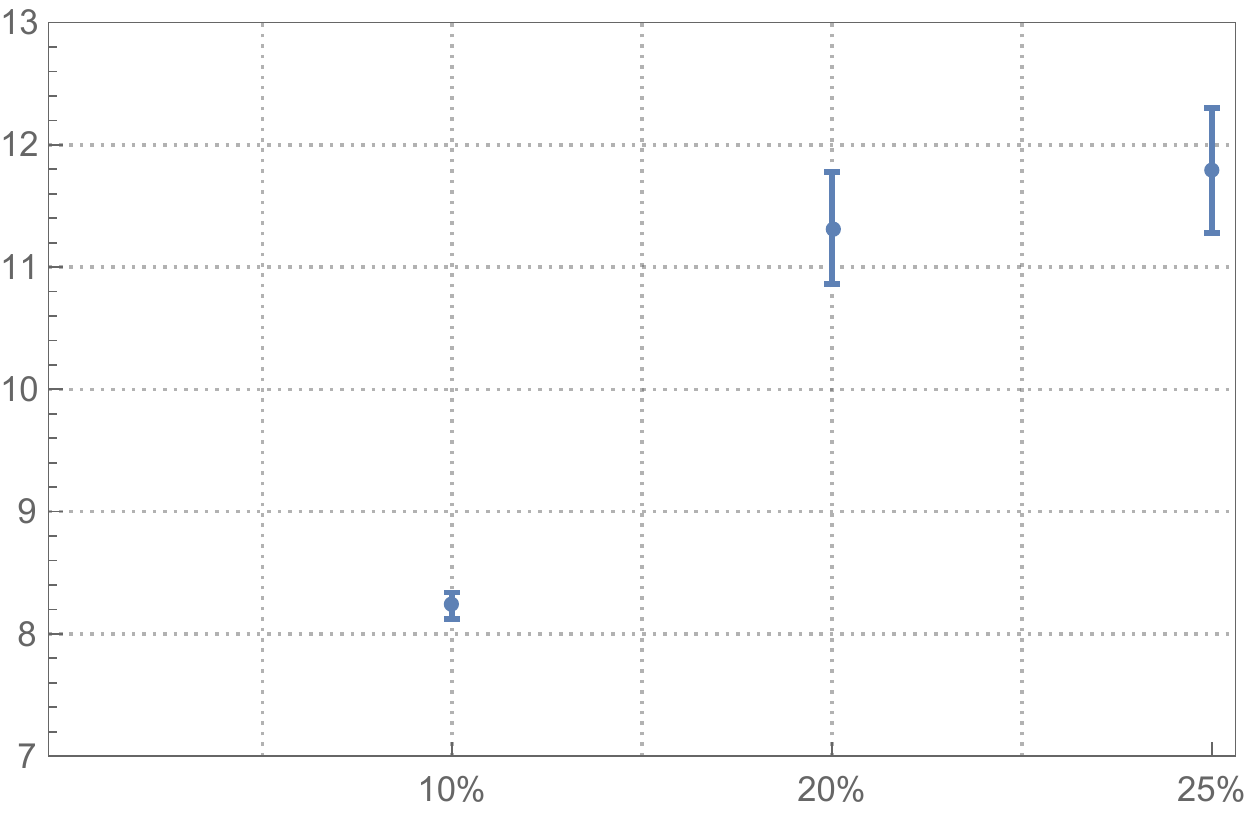}}} %
  \caption{Important Factors of PRIHGA with respect to RPD}
    \label{fig:A2cal}
\end{figure}

The Figure \ref{fig:A2cal} presents the 95\% confidence level error bounds on the means of the four factors of PRIHGA that all are found to be significantly important. The most important factor is the population size this time (F-value=25.66). The trend clearly suggests of large populations; following this, the population size parameter is tuned to 100. Secondly, SAT hybridization level in crossover is the most significant (F-value=10.98). The trend is not clear in this respect; the algorithm is tuned to have no SAT hybridization level. Here, unlike in PARHGA, the high SAT hybridization level preference does not seem existent. Thirdly, the elitist probability seems significant (F-value=10.36); this is tuned to the lower level of 0.6. Lastly, the remaining factor of selection and migration percentages also shows statistical significance level (F-value=11.10). PRIHGA is tuned to have 10\% selection and 10\% migration operators, leaving 80\% room for offspring solutions in each generation.    

\begin{figure}[H]
    \centering
    \includegraphics[scale=0.5]{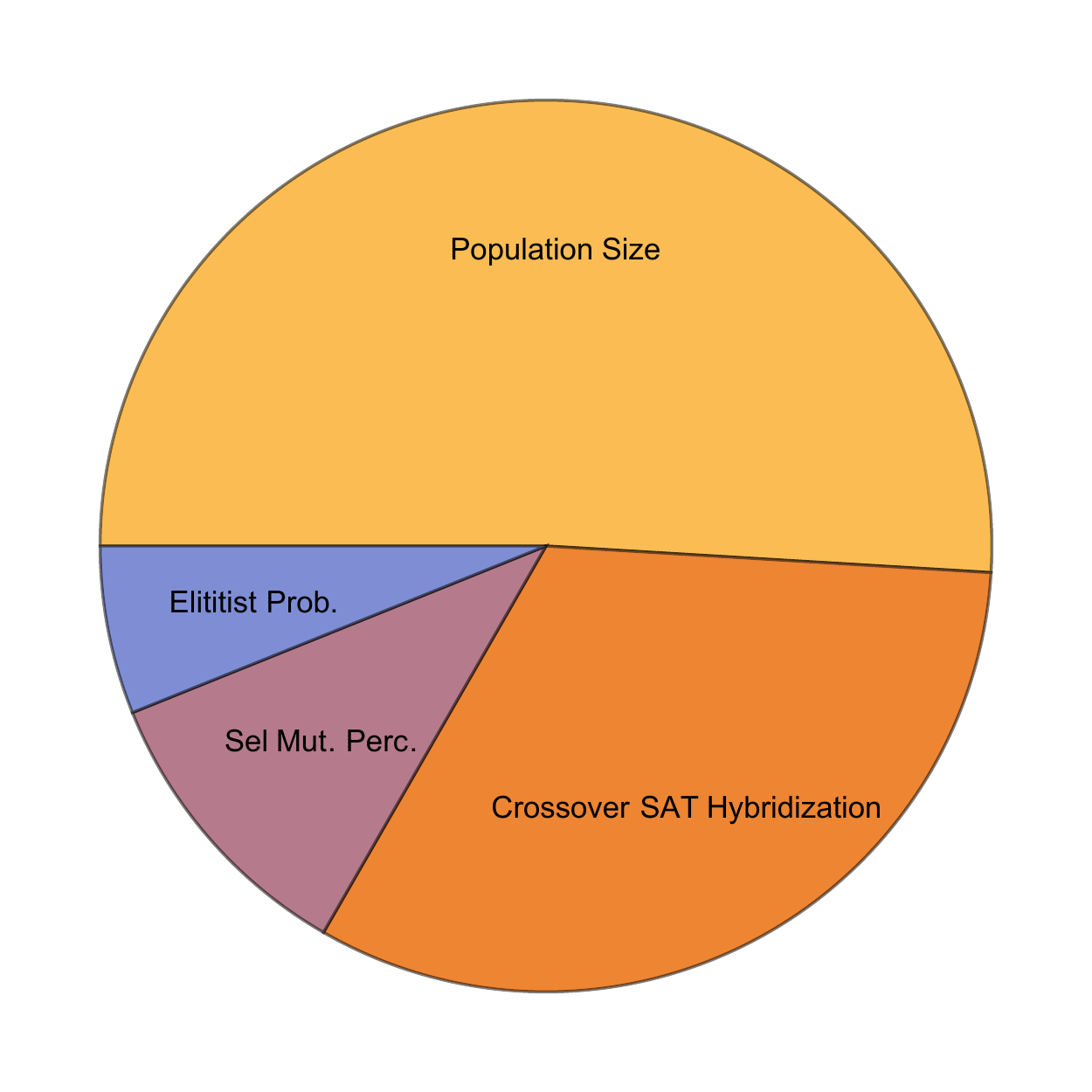}
    \caption{Pie Chart Output of the Perceptron Factor Importance Analysis for PRIHGA}
    \label{fig:percept22}
\end{figure}

Again, factor importance analysis of multi-layer perceptrons is used here for having additional opinion on the significance levels of the factors. The same setting described in Section \ref{sec:cal_1} is also used for PRIHGA. This time the output of this analysis not completely but almost matches with ANOVA. According to the sensitivity of the perceptron model, differently from ANOVA, the elitist probability factor is less important than the selection and migration percentages in the generations. The importance of the factors by the perceptron tool is illustrated with a pie chart in Figure \ref{fig:percept22}.

\subsection{Benchmarking Hybrid Genetic Algorithms}

The calibrated two hybrid genetic algorithms are tested on the 12 instances of the Toronto benchmark instances (skipping the very large Pur-s-93 instance due to practicality) in this section. Each run of PARHGA and PRIHGA is limited by the time limit of 5 hours and 3 days, respectively. These limits are selected such that both algorithms are able to produce a good number of generations for all 12 instances. PRIHGA needs significantly more time than PARGHA, due to the differences in the generation updates of the two algorithms. While PARHGA only generates one offspring in each generation, PRIHGA updates many solutions in its population in each iteration. Moreover, the calibrated settings of the population sizes of the algorithms are quite different; PRIHGA is calibrated to use a population of size 100, while the population in PARHGA is calibrated to include only 20 solutions. For PARHGA, 20 runs are completed for each instance, while for PRIHGA only 10 runs are completed, due to the lengthy duration required for each run. Lastly, in all experiments here the non-parametric test of Mann-Whitney U test is used for determining significant differences between two methods per instance.   

\begin{table}[H]
    \centering
    \begin{tabular}{l|l|l|l}
    \hline
     Instances & PARHGA (ave) &  & PRIHGA (ave) \\
    \hline 
        Car-f-92 &  4.51 & \textbf{\textless} & 4.38  \\
        Car-s-91 & 5.43 & \textbf{\textless} & 5.23 \\
        Ear-f-83 & 35.62 &  no sig. & 35.36 \\
        Hec-s-92 & 10.59 & no sig. &  10.65 \\
        Kfu-s-93 &  14.19  & \textbf{\textgreater} & 14.04\\
        Lse-f-91 &  11.58 & \textbf{\textless} & 11.12 \\
        Rye-s-93 &  9.53 &  \textbf{\textless} & 9.3 \\
        Sta-f-83 &  157.12 & \textbf{\textless}  & 157.07 \\
        Tre-s-92 &  8.71 & \textbf{\textless} &  8.56 \\
        Uta-s-92 &  3.69 & \textbf{\textless}  & 3.57 \\
        Ute-s-92 &  25.51 & \textbf{\textgreater} & 25.72 \\
        Yor-f-83 &  37.3 & \textbf{\textgreater} & 37.72 \\
        \hline
        Sum & 323.78 &   & 322.72 \\
       
    \end{tabular}
       \caption{PARHGA versus PRIHGA (A \textgreater(\textless) B means A is significantly better (worse) than B)}
    \label{tab:BothGAs}
\end{table}

Firstly, the success of two hybrid genetic algorithms are compared. The Table \ref{tab:BothGAs} shows the averages of the solutions found by both algorithms. In 7 out of 12 instances, the priority based random key hybrid genetic algorithm gives significantly better solutions than the hybrid algorithm that uses partition based representation and crossover. The last line in Table \ref{tab:BothGAs} gives the total sum of the average proximity costs of all instances by two algorithms. According to this measure, these two algorithms are very much comparable. This is indeed very interesting considering that the algorithms are characteristically very different, in terms of their genetic structures, however using the same local search methods.    

For the purposes of investigating the contribution of genetic structures in the local search hybridized genetic algorithms, a multi-start local search approach is used here, named MULTLS. This approach uses the two local search methods together that are hybridized in the genetic algorithms, VDLS and HHLS, on the solutions generated by SAT-MIN heuristic, the same heuristic that is used to generate initial solutions for the genetic algorithms. Namely, the hybrid genetic algorithms are merely the genetic versions of this multi-start approach. This approach is run for 20 times for each instance and each run is limited by the time limit of 5 hours, the same time limit that is given to PARGHA. The averages of
the solutions found in the runs by two hybrid genetic algorithms and MULTLS are compared in the Table \ref{tab:GAs_vs_LS}, again with non-parametric pairwise tests. The priority based hybrid algorithm is capable of finding better solutions than MULTLS in all instances. However, the difference between PARGHA and MULTLS is not that clear, only in 6 instances the partition based hybrid genetic algorithm can provide significantly better solutions than its multi-start local search counterpart which does not posses any genetic structures, such as keeping a population of solutions and using the solutions in the population to generate new solutions. However, when the overall measure of sum of costs over all instances is considered, that is given in the last line of Table \ref{tab:GAs_vs_LS}, both hybrid algorithms seem to be superior to MULTLS approach. 


\begin{table}[H]
    \centering
    \begin{tabular}{l|l|l|l|l|l}
    \hline
     Instances & PARHGA (ave) &  & MULTLS (ave)  &   & PRIHGA (ave) \\
    \hline 
        Car-f-92 &  4.51 &    \textbf{\textless} & 4.47  &  \textbf{\textless} & 4.38  \\
        Car-s-91 & 5.43 & \textbf{\textless} & 5.34 & \textbf{\textless}&  5.23 \\
        Ear-f-83 & 35.62 &  \textbf{\textgreater}  &36.28  &  \textbf{\textless} & 35.36 \\
        Hec-s-92 & 10.59 &   \textbf{\textgreater}  & 10.88 &  \textbf{\textless}& 10.65 \\
        Kfu-s-93 &  14.19  & \textbf{\textgreater} & 14.38 & \textbf{\textless} & 14.04\\
        Lse-f-91 &  11.58 & no sig.  & 11.53 & \textbf{\textless} & 11.12 \\
        Rye-s-93 &  9.53 &  no sig. & 9.53 & \textbf{\textless} & 9.3 \\
        Sta-f-83 &  157.12 & \textbf{\textgreater} & 157.15 & \textbf{\textless}  & 157.07 \\
        Tre-s-92 &  8.71 & no sig. & 8.72&  \textbf{\textless} &  8.56 \\
        Uta-s-92 &  3.69 & \textbf{\textless} & 3.64 & \textbf{\textless} & 3.57 \\
        Ute-s-92 &  25.51 & \textbf{\textgreater} & 26.08 & \textbf{\textless} & 25.72 \\
        Yor-f-83 &  37.3 & \textbf{\textgreater} & 38.45& \textbf{\textless}& 37.72 \\
        \hline
        Sum & 323.78 &  & 326.45 & & 322.72 \\
       
    \end{tabular}
       \caption{PARHGA and PRIHGA versus MULTLS (A \textgreater(\textless) B means A is significantly better (worse) than B)}
    \label{tab:GAs_vs_LS}
\end{table}

Next, the pure genetic versions of the both local search hybridized algorithms are compared against the hybrid algorithms, with the purpose of investigating the success of local search hybridization. In the experiments of the pure genetic algorithms, only the solutions in the initial population are applied local search, only VDLS method, after the construction phase with SAT-MIN heuristic. Both algorithms are given a time limit of 5 hours per run and for each instance with each algorithm is run for 10 times. However, both of the two pure genetic algorithms are not able to find a better solution than the solutions produced in the initial population that use VDLS local search improvement in all instances. This indicates that the local search phase in the hybrid genetic algorithms is an essential component for the success of the genetic algorithms proposed here.

Now, the competitiveness of the hybrid algorithms are to be compared against the state-of-the-art approaches in the literature. Specifically, the interest is to investigate the performances of the algorithms in terms of their ability to provide good upper bounds for the benchmark instances. Firstly, the investigations direct towards the genetic type algorithms, the ones that are benchmarked on the uncapacitated Toronto instances. However, many of the genetic type algorithms, including memetic algorithms, proposed for the examination timetabling problem are tested on different instances with different constraints (see \cite{ERew}) such that only two studies in the literature are found relevant for the comparison purposes. The first study is by Pillay and Banzhaf \cite{GAinET6} which takes a two-phase approach in composing a solution for the uncapacitated examination timetabling problem. Their approach firstly handles the hard constraints and later the first-phase solution is improved for the proximity costs in the second phase. The second study is by Cote et al. \cite{Cote} which, similarly to the algorithms proposed here, also proposes a local search hybridized genetic algorithm for the examination timetabling problem. In Table \ref{tab:GAs_vs_otherGAs}, the best solutions by two hybrid genetic algorithms and also by the multi-start local search are compared against these two related genetic type methods on the 12 instances of the Toronto data set. The both of the hybrid genetic algorithms proposed here can provide bounds as good as the two existing genetic type methods in the literature. Moreover, for two of the instances, the proposed hybrid algorithms are able to find better bounds than the existing genetic approaches. Again, for the performance comparisons on the overall data set, the last line in Table \ref{tab:GAs_vs_otherGAs} gives the sum of the costs over all instances tested. According to this measure, the two hybrid algorithms proposed here and the hybrid algorithm proposed by Cote et al. \cite{Cote} are very much comparable. However, the multi-start local search approach used here also seems competitive against the two-phase genetic approach proposed by Pillay and Banzhaf \cite{GAinET6}. In four instances, MULTLS can find better solutions than Pillay and Banzhaf's algorithm and in one case it can also find a better solution than Cote et al.'s hybrid genetic algorithm.          

The Table \ref{tab:Bench} gives the best bounds found for the 12 instances of Toronto uncapacitated instances by the well-performing non-genetic type methods in the literature. The last line in the table shows the sum of the costs over considered 12 instances. The hybrid genetic algorithms proposed here seem competitive against these approaches, however, not being able to produce better bounds. 


\begin{table}[H]
    \centering
     \resizebox{0.7\textwidth}{!}{\begin{minipage}{\textwidth}
    \begin{tabular}{l|l|l|l|l|l}
    \hline
     Instances & PARHGA (best) &  PRIHGA (best) & MULTLS (best) & PIllay 2010 (best) & Cote 2004 (best)\\
    \hline 
        Car-f-92 &  4.40 & 4.34 & 4.47& \textbf{4.2} & \textbf{4.2}  \\
        Car-s-91 &  5.36 & 5.14 & 5.33 & \textbf{4.9} & 5.4 \\
        Ear-f-83 & 34.93 & 34.93 & 36.28 & 35.9 & \textbf{34.2} \\
        Hec-s-92 & 10.48 & 10.52 & 10.88 & 11.5 & \textbf{10.4} \\
        Kfu-s-93 &  14.03 & \textbf{13.86} & 14.38 & 14.4 & 14.3 \\
        Lse-f-91 &  11.38 & 10.97 & 11.53 & \textbf{10.9} & 11.3 \\
        Rye-s-93 &  9.29 & 9.24 & 9.53 & 9.3 & \textbf{8.8}  \\
        Sta-f-83 &  157.05 & 157.05 & 157.15 & 157.8& \textbf{157} \\
        Tre-s-92 &  8.60 & 8.51 & 8.72 & \textbf{8.4} & 8.6 \\
        Uta-s-92 &  3.61 & 3.56 & 3.64 & 3.4 & \textbf{3.5} \\
        Ute-s-92 &  \textbf{25.29} & 25.51 & 26.08 & 27.2 & 25.3 \\
        Yor-f-83 &  36.67 & 37.56 & 38.45 & 39.3 & \textbf{36.4} \\
        \hline
        Sum & 321.09 & 321.19 &  326.44 & 327.2 & 319.4 \\
       
    \end{tabular}
      \end{minipage}}
       \caption{Best Solutions by PARHGA, PRIHGA and MULTLS Compared to Other Genetic Methods}
    \label{tab:GAs_vs_otherGAs}
\end{table}

\begin{table}[H]
    \resizebox{0.7\textwidth}{!}{\begin{minipage}{\textwidth}
    \begin{tabular}{l|l|l|l|l|l|l}
    \hline
     Instances & Carter96  & Caramia08  & Burke08  & Demeester12  &  Alzaqebah15  & Leite18 \\
     & \cite{Carter} &  \cite{Caramia} & \cite{Bykov} & \cite{Demeester} & \cite{Bees} & \cite{GAinET5} \\
    \hline 
        Car-f-92 &  6.2 & 6 & 3.81 & 3.78 & 3.88 & 3.68 \\
        Car-s-91 &  7.1 & 6.6 & 4.58 & 4.52 & 4.38 & 4.31 \\
        Ear-f-83 & 36.4 & 29.3 & 32.65 & 32.49 & 33.34 & 32.48 \\
        Hec-s-92 & 10.8 & 9.2 & 10.06 & 10.03 & 10.39 & 10.03 \\
        Kfu-s-93 &  14 & 13.8 & 12.81 & 12.9 & 13.23 & 12.81 \\
        Lse-f-91 &  10.5 & 9.6 & 9.86 & 10.04 & 10.52 & 9.78 \\
        Rye-s-93 &  7.3 & 6.8 & 7.93 & 8.05 & 8.92 & 7.89 \\
        Sta-f-83 &  161.5 & 158.2 & 157.03 & 157.03 & 157.06 & 157.03 \\
        Tre-s-92 &  9.6 & 9.4 & 7.72 & 7.69 & 7.89 & 7.66 \\
        Uta-s-92 &  3.5 & 3.5 & 3.16 & 3.13 & 3.13 & 3.01 \\
        Ute-s-92 &  25.8 & 24.4 & 24.79 & 24.77 & 25.12 & 24.80\\
        Yor-f-83 &  41.7 & 36.2 & 34.78 & 34.64 & 35.49 & 34.45\\
        \hline
        Sum & 334.4 & 313 & 309.18 & 309.07 & 313.35 & 307.93 \\
       
    \end{tabular}
    
    \end{minipage}}
       \caption{Best Solutions by Non-genetic Type State-of-the-art Methods}
    \label{tab:Bench}
\end{table}

\section{Conclusions and Future Research Directions}\label{sec:4}

This paper aims to investigate the success of local search hybridized genetic algorithms with indirect representations in solving the examination timetabling problem. For this purpose, two structurally different genetic algorithms are devised. These algorithms are inspired from the successful genetic algorithms proposed for different but similar problems in the literature. Specifically, the inspiration source of the first algorithm is a partition based hybrid genetic algorithm proposed for graph coloring problems and for the second algorithm, a priority based random key genetic algorithm that is proposed for a project scheduling problem is the source of inspiration. One computationally light and one intensive local search methods are proposed and used in the hybrid algorithms. The calibration experiments indicate that although the intensive local search method increases the computational burden, the quality of the solutions are significantly better. For this reason, it might be the case that even more intense and carefully-developed local search methods would offer more benefits to the success of hybrid genetic algorithms. The crossovers used in the algorithms are parameterized with saturation degree heuristic to experiment on the potential of using lightened crossovers. The use of light crossovers shows some indications of significant benefits in one of the hybrid algorithms. Future research might delve into developing hybrid light crossovers. Also, these type of crossovers may have potential to be useful for graph coloring problems, considering their advantage on producing non-conflicting solutions. The initial solutions of the algorithms are generated by two saturation degree heuristics. These heuristics differ only in their assignment rules. The distance based assignment rule is newly proposed here and it shows significant superiority over the regular minimum assignment rule. 

The hybridization of local search with both of the genetic algorithms seem successful. The experiments on the pure genetic algorithms reveal that the local search component is very vital in the performances of the algorithms such that without it, with only genetic structures, the algorithms can not be successful. This confirms the conclusion reported in the earlier studies on the success of pure genetic algorithms for the examination timetabling problem. Also, the experiments on the multi-start local search method indicate that the genetic structures are contributing to the performances of the algorithms. 

The experiments on the both hybrid algorithms reveal that the performances of the algorithms are very similar to each other and also similar to other genetic-type algorithms proposed already in the literature, on the uncapacitated Toronto instances. The performances by genetic-type algorithms, including the two hybrid algorithms proposed here, seem not as good as the non-genetic type state-of-the-art heuristics. However, it can not be claimed that in general, the local search hybridized population based heuristics are not worthy of attention. For instance, the cellular memetic algorithm proposed in a very recent study by Leite et al. \cite{GAinET5} is one of the most successful algorithms proposed for the uncapacitated examination timetabling up to now. The main future research interest is to investigate the potential of this type of local search hybridized population search methods. 




\bibliography{ref}

\begin{thebibliography}{}

\bibitem[Akkan and Gulcu, 2018]{GAinET2}
Akkan, C. and Gulcu, A. (2018).
\newblock A bi-criteria hybrid genetic algorithm with robustness objective for
  the course timetabling problem.
\newblock {\em Computers and Operations Research}, 90:22--32.

\bibitem[Alzaqebah and Abdullah, 2015]{Bees}
Alzaqebah, M. and Abdullah, S. (2015).
\newblock Hybrid bee colony optimization for examination timetabling problems.
\newblock {\em Computers and Operations Research}, 54:142--154.

\bibitem[Brelaz, 1979]{SAT}
Brelaz, D. (1979).
\newblock New methods to color the vertices of a graph.
\newblock {\em Communications of the ACM}, 22:251--256.

\bibitem[Burke et~al., 2010]{VNS}
Burke, E., Eckersley, A., McCollum, B., Petrovic, S., and Qu, R. (2010).
\newblock Hybrid variable neighbourhood approaches to university exam
  timetabling.
\newblock {\em European Journal of Operational Research}, 206:46--53.

\bibitem[Burke and Bykov, 2008]{Bykov}
Burke, E.~K. and Bykov, Y. (2008).
\newblock {\em A late acceptance strategy in hill-climbing for exam timetabling
  problems. In: Proceedings of the Seventh International Conference on the
  Practice and Theory of Automated Timetabling, PATAT 2008}.
\newblock PATAT 2008.

\bibitem[Burke et~al., 2009]{ClassificationHH}
Burke, E.~K., Hyde, M., Kendall, G., Ochoa, G., Ozcan, E., and Woodward, J.
  (2009).
\newblock A classification of hyper-heuristic approaches.
\newblock Technical report, University of Nottingham.

\bibitem[Caramia et~al., 2008]{Caramia}
Caramia, M., Dell'Olmo, P., and Italiano, G. (2008).
\newblock Novel local-search-based approaches to university examination
  timetabling.
\newblock {\em INFORMS Journal on Computing}, 20:86--99.

\bibitem[Carter et~al., 1996]{Carter}
Carter, M., Laporte, G., and Lee, S. (1996).
\newblock Examination timetabling: algorithmic strategies and applications.
\newblock {\em The Journal of the Operational Research Society}, 47:373--383.

\bibitem[Cheong et~al., 2009]{GAinET1}
Cheong, C., Tan, K., and Veeravalli, B. (2009).
\newblock A multi-objective evolutionary algorithm for examination timetabling.
\newblock {\em Journal of Scheduling}, 12:121–146.

\bibitem[Corne et~al., 1994]{GAinETs1}
Corne, D., Ross, P., and Fang, H. (1994).
\newblock Evolutionary timetabling: Practice, prospects and work in progress.
  in p. prosser (ed.),.
\newblock {\em Proceedings of UK planning and scheduling SIG workshop}.

\bibitem[Cote et~al., 2004]{Cote}
Cote, P., Wong, T., and Sabourin, R. (2004).
\newblock {\em Application of a hybrid multi-objective evolutionary algorithm
  to the uncapacitated exam proximity problem, in: E.K. Burke, M. Trick (Eds.),
  Practice and Theory of Timetabling V, 5th International Conference, PATAT
  2004}, pages 294--312.
\newblock Springer, Berlin, Heidelberg.

\bibitem[Demeester et~al., 2012]{Demeester}
Demeester, P., Bilgin, B., De~Causmaecker, P., and Vanden~Berghe, G. (2012).
\newblock A hyperheuristic approach to examination timetabling problems:
  benchmarks and a new problem from practice.
\newblock {\em Journal of Scheduling}, 15:83--103.

\bibitem[Galinier and Hao, 1999]{GAGC1}
Galinier, P. and Hao, J. (1999).
\newblock Hybrid evolutionary algorithms for graph coloring.
\newblock {\em Journal of Combinatorial Optimization}, 3:379--397.

\bibitem[Glass and Prugel-Bennett, 2003]{GAGC1follow}
Glass, C. and Prugel-Bennett, A. (2003).
\newblock Genetic algorithm for graph coloring: exploration of {Galinier} and
  {Hao's} algorithm.
\newblock {\em Journal of Combinatorial Optimization}, 7:229--236.

\bibitem[Holland, 1992]{GAHist}
Holland, J. (1992).
\newblock {\em Adaptation in aatural and artificial Systems: an introductory
  analysis with applications to biology, control and artificial intelligence}.
\newblock MIT Press Cambridge.

\bibitem[Lei and Shi, 2017a]{GAinET4}
Lei, Y. and Shi, J. (2017a).
\newblock A memetic algorithm based on {MOEA/D} for the examination timetabling
  problem.
\newblock {\em Soft Computing}, 22:1511--1523.

\bibitem[Lei and Shi, 2017b]{GAinET3}
Lei, Y. and Shi, J. (2017b).
\newblock A {NNIA} scheme for timetabling problems.
\newblock {\em Journal of Optimization}, page~11.

\bibitem[Leite et~al., 2018]{GAinET5}
Leite, N., Fernandes, C., Melicio, F., and Rosa, A. (2018).
\newblock A cellular memetic algorithm for the examination timetabling problem.
\newblock {\em Computers and Operations Research}, 94:118--138.

\bibitem[Mendes et~al., 2009]{GA2}
Mendes, J., Goncalves, J., and Resende, M. (2009).
\newblock A random key based genetic algorithm for the resource constrained
  project scheduling problem.
\newblock {\em Computers and Operations Research}, 36:92--109.

\bibitem[Pillay, 2016]{HH}
Pillay, N. (2016).
\newblock A review of hyper-heuristics for educational timetabling.
\newblock {\em Annals of Operations Research}, 239:3--38.

\bibitem[Pillay and Banzhaf, 2010]{GAinET6}
Pillay, N. and Banzhaf, W. (2010).
\newblock An informed genetic algorithm for the examination timetabling
  problem.
\newblock {\em Applied Soft Computing}, 10:457--467.

\bibitem[Qu et~al., 2008]{ERew}
Qu, R., Burke, E.~K., McCollum, B., Merlot, L., and Lee, S. (2008).
\newblock A survey of search methodologies and automated system development for
  examination timetabling.
\newblock {\em Journal of Scheduling}, 12:55--89.

\bibitem[Ross et~al., 1998]{GAinETs2}
Ross, P., Hart, E., and Corne, D. (1998).
\newblock Some observations about {GA}-based exam timetabling. in e. k. burke
  \& m. w. carter (eds.).
\newblock {\em Lecture notes in computer science: Practice and theory of
  automated timetabling II: selected papers from the 2nd international
  conference}, 1408:115--129.

\bibitem[Ruiz et~al., 2006]{GAexper}
Ruiz, R., Maroto, C., and Alcaraz, J. (2006).
\newblock Two new robust genetic algorithms for the flowshop scheduling
  problem.
\newblock {\em Omega}, 34:461--476.

\bibitem[Thompson and Dowsland, 1996]{Kempe}
Thompson, J. and Dowsland, K. (1996).
\newblock Variants of simulated annealing for the examination timetabling
  problem.
\newblock {\em Annals of Operations Research}, 63:105--128.

\bibitem[Turabieh and Abdullah, 2011]{EMmethod}
Turabieh, H. and Abdullah, S. (2011).
\newblock An integrated hybrid approach to the examination timetabling problem.
\newblock {\em Omega}, 39:598--607.

\end{thebibliography}

\bibliographystyle{apalike} 
\end{document}